\documentclass[conference]{IEEEtran}
\IEEEoverridecommandlockouts
\usepackage{cite}
\usepackage{tikz}
\usetikzlibrary{positioning}
\usepackage{amsmath,amssymb,amsfonts}
\usepackage{algorithmic}
\usepackage{graphicx}
\usepackage{textcomp}
\usepackage{pifont}
\usepackage{comment}
\usepackage{xcolor}
\usepackage{balance}
\usepackage{mathtools}

\usepackage{caption}
\usepackage{etoolbox}
\newcommand{\insertfig}{
    \setcounter{figure}{0}
    \vspace{1em} 
    \begin{center}
        \setlength{\fboxsep}{0pt}
        \captionsetup{type=figure} 
        \includegraphics[width=\linewidth]{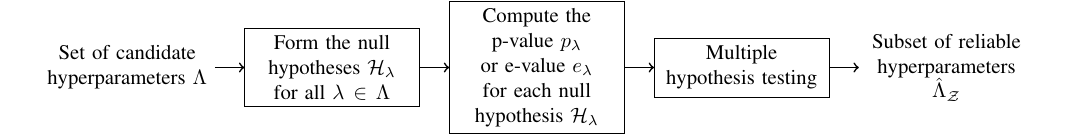}
        \caption{Block diagram for LTT-based hyperparameter selection methods.}
    \label{fig:block_diagram}
    \end{center}
    \vspace{0.5em} 
}

\makeatletter
\apptocmd{\@maketitle}{\centering\insertfig}{}{}
\makeatother

\def\BibTeX{{\rm B\kern-.05em{\sc i\kern-.025em b}\kern-.08em
    T\kern-.1667em\lower.7ex\hbox{E}\kern-.125emX}}
\begin{document}

\title{Ensuring Reliability via Hyperparameter Selection: Review and Advances\\

\thanks{This work was supported by the European Union’s Horizon Europe project CENTRIC (101096379), by the Open Fellowships of the EPSRC (EP/W024101/1), and by the EPSRC project (EP/X011852/1).}
}

\author{\IEEEauthorblockN{Amirmohammad Farzaneh and Osvaldo Simeone}  
\IEEEauthorblockA{\textit{Centre for Intelligent Information Processing Systems,}\\
\textit{Department of Engineering,} \\  
\textit{King's College London,}\\
\textit{London, United Kingdom} \\  
\{amirmohammad.farzaneh, osvaldo.simeone\}@kcl.ac.uk}  
}

\maketitle

\begin{abstract}
Hyperparameter selection is a critical step in the deployment of artificial intelligence (AI) models, particularly in the current era of foundational, pre-trained, models. By framing hyperparameter selection as a multiple hypothesis testing problem, recent research has shown that it is possible to provide statistical  guarantees on population risk measures attained by the selected hyperparameter. This paper reviews the Learn-Then-Test (LTT) framework, which formalizes this approach, and explores several extensions tailored to engineering-relevant scenarios. These extensions encompass different risk measures and statistical guarantees,  multi-objective optimization, the incorporation of prior knowledge and dependency structures into the hyperparameter selection process, as well as adaptivity. The paper also includes illustrative applications for communication systems.
\end{abstract}


\section{Introduction}

\begin{table*}[!t]
\caption{Comparison of LTT-based hyperparameter selection methods}
\begin{center}
\begin{tabular}{|c||c|c|c|c|c|}
\hline
\textbf{Scheme} & \textbf{Risk Measure}& \textbf{Statistical Guarantee}& \textbf{Multi-Objective} & \textbf{Side Information} & \textbf{Adaptive} \\
\hline
LTT \cite{angelopoulos2021learn} & Average & FWER &  \ding{55} & \ding{55}& \ding{55} \\
\hline
QLTT \cite{farzaneh2024quantile}& Quantile & FWER & \ding{55} & \ding{55}& \ding{55} \\
\hline
IB-MHT \cite{farzaneh2024statistically}& Mutual Information & FWER & \ding{55} & \ding{55}& \ding{55} \\
\hline
PT \cite{laufer2022efficiently}& Average & FWER & \ding{51} & \ding{55} & \ding{55}\\
\hline
RG-PT \cite{farzaneh2025multi}& Average & FDR & \ding{51} & \ding{51} & \ding{55}\\
\hline
aLTT \cite{zecchin2024adaptive}& Average & FWER & \ding{55} & \ding{55} & \ding{51}\\
\hline
\end{tabular}
\label{tab1}
\end{center}
\end{table*}

The selection of hyperparameters is a fundamental challenge in the optimization of machine learning models, directly impacting efficiency and generalization \cite{yang2020hyperparameter}. Traditional approaches rely on heuristic search strategies, such as grid search and Bayesian optimization, which prioritize empirical performance but fail to provide formal statistical guarantees \cite{bergstra2012random, snoek2012practical}. The lack of performance guarantees may prevent the use of AI models in sensitive domains, such as healthcare \cite{dzau2023achieving} and engineering \cite{bensalem2023indeed}. Recently, a growing body of research has explored hyperparameter selection from a statistical inference perspective, treating it as a multiple hypothesis testing (MHT) problem. This paradigm provides a principled way to control error rates in the hyperparameter tuning process, making it possible to select hyperparameters in a manner that ensures formal reliability targets.
The Learn-Then-Test (LTT) framework \cite{angelopoulos2021learn} formalizes this perspective by structuring hyperparameter selection as a two-stage process. As illustrated in Figure \ref{fig:block_diagram}, this consists of the following phases: (\textit{i}) Determine a set $\Lambda$ of candidate hyperparameters using any methodology; (\textit{ii}) Test the performance of the candidate hyperparameters via MHT to determine a subset $\hat{\Lambda}_\mathcal{Z}\subseteq \Lambda$ of hyperparameters with formal statistical performance guarantees.  LTT has been applied to a number of settings, including language models \cite{schuster2022confident} and vision models \cite{angelopoulos2022image}.

LTT focuses on the control of average risk measures by using standard MHT methods that target the \textit{family-wise error rate} (FWER) criterion. Accordingly, the goal is to ensure that the produced subset $\hat{\Lambda}_\mathcal{Z}$ of hyperparameters include no unreliable configurations with sufficiently high probability. Beyond this standard LTT formulation, real-world engineering applications introduce additional complexities, such as the need to account for risk-aware performance measures \cite{karasik2022learning, farzaneh2024quantile}, the freedom to explore multi-objective trade-offs \cite{sun2015multi}, the presence of prior domain knowledge \cite{raviv2024adaptive}, and the cost of acquiring data \cite{hou2025if}. As summarized in Table \ref{tab1}, addressing these aspects requires extensions to the LTT framework, which are reviewed in this tutorial-style paper.

Specifically, Section \ref{sec:LTT} provides a review of the fundamental LTT framework. Each subsequent section explores one of the aspects described by the columns of Table \ref{tab1}. Specifically, Section \ref{sec:risk_measures} explores different risk measures; Section \ref{sec:statistical} covers statistical guarantees; Section \ref{sec:multiple} discusses multi-objective optimization criteria; the integration of side information is covered in Section \ref{sec:prior}; and adaptive LTT-based methods are reviewed in Section \ref{sec:adaptive}. Conclusions and future research directions are presented in Section \ref{sec:conclusion}.

\section{LTT-Based Hyperparameter Selection}
\label{sec:LTT}

The goal of LTT-based hyperparameter selection methods is to control a risk measure $R(\lambda)$ by choosing a vector of hyperparameters $\lambda$. As detailed in Section \ref{sec:risk_measures}, the risk measure is a general functional of the true, unknown, distribution $P_Z$ of the data $Z\sim P_Z$. LTT-based schemes require the following inputs:
\begin{enumerate}
    \item A discrete set $\Lambda$ of candidate hyperparameter configurations $\lambda$.
    \item A held-out calibration data set $\mathcal{Z} = \{Z_i\}_{i = 1}^{|\mathcal{Z}|}$, with i.i.d. data points $Z_i\sim P_Z$.
\end{enumerate}

LTT-based methods associate with each hyperparameter $\lambda \in \Lambda$ the null hypothesis $\mathcal{H}_\lambda$ that hyperparameter $\lambda$ is unreliable. Formally, the null hypothesis $\mathcal{H}_\lambda$ is defined as
\begin{equation}
\label{eq:null_hypothesis}
    \mathcal{H}_\lambda:\quad R(\lambda)<\alpha,
\end{equation}
where $\alpha$ is a user-specified threshold. Rejecting the null hypothesis $\mathcal{H}_\lambda$ is equivalent to deeming $\lambda$ reliable.

Testing is performed by evaluating, based on the calibration data $\mathcal{Z}$, a valid p-value $p_\lambda$, or e-value $e_\lambda$ \cite{ramdas2024hypothesis}, for each null hypothesis $\mathcal{H}_\lambda$. P-values and e-values are test statistics that provide measures of evidence in favor or against the null hypothesis. While p-values are commonly used for fixed-sample hypothesis testing, e-values allow for more flexible testing strategies. Notably, they support sequential methods whereby one continuously monitors the testing outcomes, stopping whenever sufficient evidence against the null hypothesis \( \mathcal{H}_\lambda \) is received \cite{ramdas2024hypothesis}.

As shown in Figure \ref{fig:block_diagram}, using the set of p-values $\{p_\lambda\}_{\lambda \in \Lambda}$ or e-values $\{e_\lambda\}_{\lambda \in \Lambda}$ for the set $\Lambda$ of candidate hyperparameters, LTT-based methods use an MHT algorithm \cite{rice2007mathematical} to test the hypotheses $\{\mathcal{H}_\lambda\}_{\lambda \in \Lambda}$. As detailed in Sec. \ref{sec:statistical}, the output of MHT is a subset $\hat{\Lambda}_\mathcal{Z} \subseteq \Lambda$ of hyperparameters with the guarantee that it contains no, or few, unreliable hyperparameters $\lambda$ violating the condition (\ref{eq:null_hypothesis}), i.e., $R(\lambda)\geq \alpha$.

\section{Risk Measures}
\label{sec:risk_measures}
Consider an instantaneous risk function \( r_\lambda(Z) \), which quantifies the risk or loss associated with a test instance \( Z \) when the model is deployed using hyperparameter \( \lambda \). To capture the overall risk across the entire data distribution, different \textit{risk measures} can be adopted. This section provides an overview of risk measures studied in the literature.

The standard risk measure used in LTT \cite{angelopoulos2021learn} is the \textit{average} risk
\begin{equation}
\label{eq:average_risk}
R_{\text{avg}}(\lambda) = \mathbb{E}_{Z}\left[r_\lambda(Z)\right],
\end{equation}
where the expectation is taken with respect to the random variable $Z\sim P_Z$. With this definition, assuming a bounded risk function $r_\lambda(Z)\in (0,1]$, a valid p-value for the null hypothesis $\mathcal{H}_\lambda$ in (\ref{eq:null_hypothesis}) can be calculated by using Hoeffding's inequality \cite{angelopoulos2021learn}
\begin{equation}
\label{eq:pval_average}
    p_\lambda^{\text{avg}}(\mathcal{Z}) = e^{-2|\mathcal{Z}|(\alpha - \hat{R}_\text{avg}(\lambda|\mathcal{Z}))^2_+},
\end{equation}
where
\begin{equation}
\label{eq:empirical_estimate}
\hat{R}_\text{avg}(\lambda|\mathcal{Z}) = \frac{1}{|\mathcal{Z}|} \sum_{Z\in \mathcal{Z}} r_\lambda(Z)
\end{equation}
is the empirical estimate for the average risk $R_\text{avg}(\lambda)$ obtained using data set $\mathcal{Z}$.

While the average risk \( R_{\text{avg}}(\lambda) \) provides a useful baseline for evaluating performance, it may fail to capture the variability of the risk across different instances. In practice, performance metrics beyond averages are often of interest, including uncertainty-aware measures \cite{abdar2021review} and information-theoretic measures \cite{rodriguez2024information}.

Among the uncertainty-aware risk measures, \textit{quantiles} play a central role. For example, in cellular wireless systems, one typically wishes to control the lower quantiles of \textit{key performance indicators} (KPIs), such as throughput and latency, to offer performance guarantees also to cell-edge users \cite{karasik2022learning}.

The \textit{$q$-quantile risk} $R_q(\lambda)$ attained by a hyperparameter $\lambda$ is defined as
\begin{equation}
\label{eq:quantile_risk}
    R_q(\lambda) = \inf_{r\geq 0} \left\{\mathrm{Pr}_{Z}[r_\lambda(Z) \leq r] \geq 1-q  \right\}.
\end{equation}
This measure evaluates the smallest risk \( r \) such that at least a fraction \( 1-q \) of the risk values \( r_\lambda(Z) \) lie below it. Thus, controlling the $q$-quantile risk, i.e., ensuring the inequality $R_q(\lambda)<\alpha$, guarantees that the worst-case risk among the top-\( q \) fraction of test instances is no smaller than $\alpha$.

\begin{figure}
    \centering
    \includegraphics[width=\columnwidth]{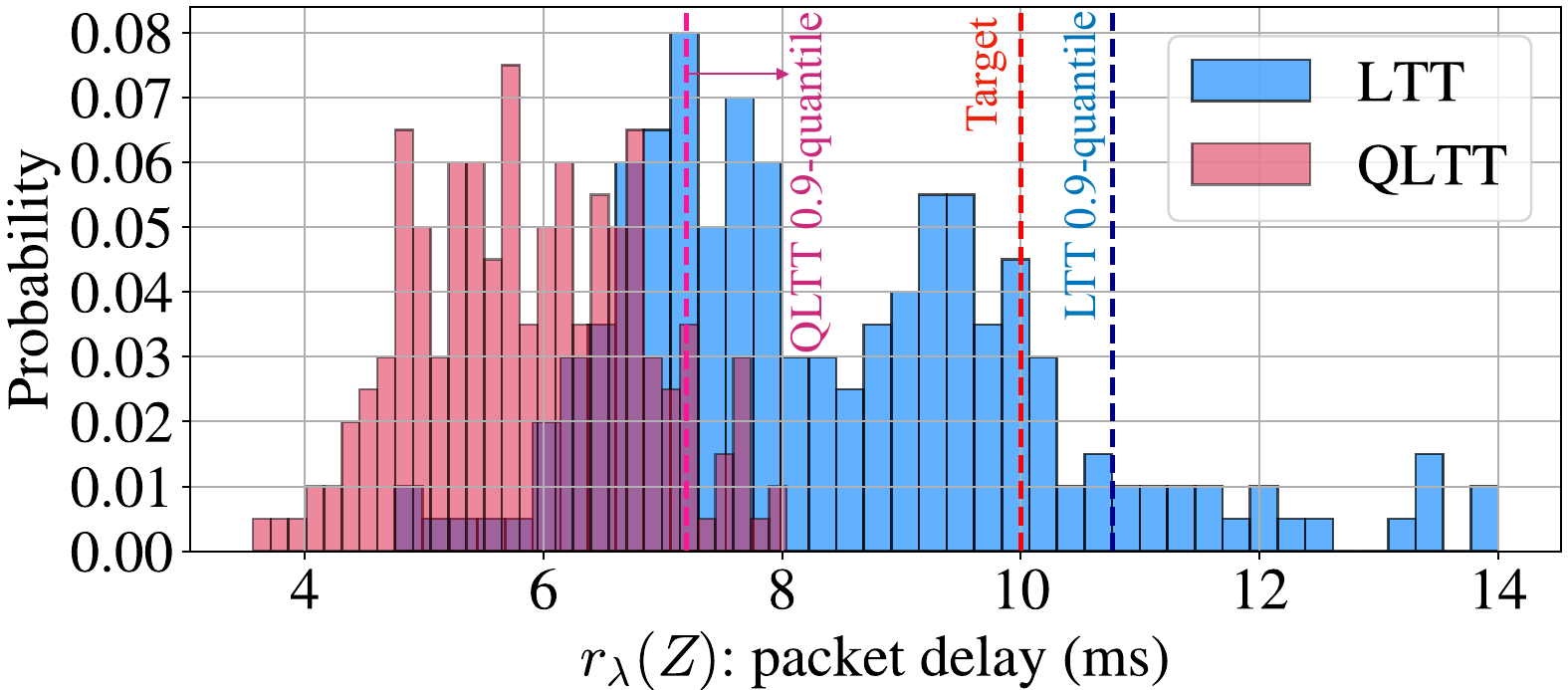}
    \caption{Packet delay distribution for LTT and QLTT in a resource allocation problem (adapted from \cite{farzaneh2024quantile} by using a different calibration data set $\mathcal{Z}$).}
    \label{fig:quantile}
\end{figure}

QLTT \cite{farzaneh2024quantile} applies the LTT procedure described in Section \ref{sec:LTT} to the control of the quantile risk $R_q(\lambda)$ via the use of the p-value in \cite{farzaneh2024quantile}. Figure \ref{fig:quantile} compares the distribution of KPIs attained in a radio resource allocation problem \cite{Nokia} by LTT and QLTT (see details in \cite{farzaneh2024quantile}). The objective here is to ensure that at least 90\% of users experience a packet delay below the 10 ms target, i.e., that the $q = 0.9$-quantile of packet delays remains under this threshold. Since LTT controls the average risk, it fails to meet the quantile requirement, whereas QLTT successfully ensures compliance with the target.

An example of an information-theoretic risk measure is \textit{mutual information} (MI). In \cite{farzaneh2024statistically}, MI-based reliability requirements were considered in the context of the information bottleneck problem \cite{tishby2000information}. In it, the hyperparameter $\lambda$ of a feature extraction function $f_\lambda$ that computes features $T_\lambda$ from covariates $X$ is optimized such that the MI between extracted features $T_\lambda$ and targets $Y$, shown with $I(T_\lambda;Y)$, is above a threshold $\alpha$. \cite{farzaneh2024statistically} calculates a valid p-value for the hypothesis $\mathcal{H}_\lambda: I(T_\lambda;Y) < \alpha$, enabling valid MHT using MI as the risk measure.

\section{Statistical Guarantees}
\label{sec:statistical}

As discussed in Section \ref{sec:LTT}, the goal of LTT-based hyperparameter selection methods is to identify a subset, $\hat{\Lambda}_\mathcal{Z}$, of hyperparameters $\lambda$ in the candidate set $\Lambda$ that satisfies the statistical guarantees on the risk measure $R(\lambda)$ (see Sec. \ref{sec:risk_measures}). As reviewed in this section, the statistical guarantees can be formalized in different ways.

LTT \cite{angelopoulos2021learn} employs the FWER as the criterion, which ensures that the probability of including \textit{any} unreliable hyperparameter in subset $\hat{\Lambda}_\mathcal{Z}$ remains below a user-specified threshold $\delta$. Formally, the selected subset $\hat{\Lambda}_\mathcal{Z}$ satisfies the guarantee
\begin{equation}
\label{eq:fwer}
\mathrm{Pr}_{\mathcal{Z}} \big[ R(\lambda) \leq \alpha \;
                                \text{for all } \lambda \in \hat{\Lambda}_\mathcal{Z} \big] 
                                \geq 1 - \delta,
\end{equation}
where the probability is evaluated with respect to the data set $\mathcal{Z}$.

As a simple method for controlling the FWER, Bonferroni correction selects the subset  
\begin{equation}
    \hat{\Lambda}_\mathcal{Z} = \left\{\lambda \in \Lambda: p_\lambda < \frac{\delta}{|\Lambda|}\right\}.
\end{equation}
However, as the size of the initial candidate set \( \Lambda \) increases, Bonferroni's method becomes overly conservative, leading to fewer selected hyperparameters. To mitigate this limitation, \textit{fixed sequence testing} (FST) \cite{bauer1991multiple} can be applied when prior knowledge is available regarding the relative reliability of different hyperparameters.

With FST, the hyperparameters \( \lambda \in \Lambda \) are first arranged in a predefined order, denoted as \( (\lambda_1, \ldots, \lambda_{|\Lambda|}) \), where \( \lambda_1 \) is presumed to be the most reliable and \( \lambda_{|\Lambda|} \) the least reliable. Then, FST sequentially evaluates each hyperparameter in descending order of expected reliability, i.e., \( \lambda_1, \lambda_2, \ldots, \lambda_{|\Lambda|} \), halting at the first instance where the inequality $p_{\lambda_j} \leq \delta$ is no longer satisfied. The final subset of reliable hyperparameters is then given by $\hat{\Lambda}_\mathcal{Z} = \left\{\lambda_1, \ldots, \lambda_j\right\}$.

As shown in \cite{einbinder2024semi}, due to the duality between confidence intervals and p-values, FST can also be applied if one has access to an upper confidence bound $R^+_\mathcal{Z}(\lambda)$ at level $\delta$ on the risk function $R(\lambda)$, satisfying the property
\begin{equation}
    \mathrm{Pr}_{\mathcal{Z}}[R(\lambda) < R^+_\mathcal{Z}(\lambda)] \geq 1-\delta.
\end{equation}
In this case, hyperparameter $\lambda$ is detected as reliable if the inequality $R^+_\mathcal{Z}(\lambda) < \alpha$ holds.

As summarized in Table \ref{tab1}, most existing LTT-based approaches rely on FWER control for statistical guarantees. However, FWER control can be excessively conservative, particularly as the number of candidate hyperparameters in set \(\Lambda\) increases. Rather than restricting the probability of including \textit{any} unreliable hyperparameters in the selected subset \(\hat{\Lambda}_\mathcal{Z}\), an alternative approach is to control the expected \textit{proportion} of unreliable hyperparameters within \(\hat{\Lambda}_\mathcal{Z}\). This criterion, known as \textit{false discovery rate} (FDR) control, ensures that the selected subset \(\hat{\Lambda}_\mathcal{Z} \subseteq \Lambda\) satisfies the inequality \cite{benjamini1995controlling}
\begin{equation}
\label{eq:FDR_statistical}
    \mathbb{E}_Z\left[\frac{\sum_{\lambda \in \hat{\Lambda}_\mathcal{Z}} \mathbf{1}\{ R(\lambda) > \alpha\}}{\max(|\hat{\Lambda}_\mathcal{Z}|, 1)}\right] \leq \delta,
\end{equation}
where \(\mathbf{1}\{\cdot\}\) is the indicator function, and the expectation is taken over the unknown data distribution \(P_Z\). FDR control is adopted by RG-PT \cite{farzaneh2025multi}, which is briefly described in Section \ref{sec:prior}.

\section{Multi-Objective Optimization}
\label{sec:multiple}

In many real-world problems, one is interested in controlling multiple risk measures, say $\{R_l(\lambda)\}_{l = 1}^L$, simultaneously. More broadly, the goal is often to address the multi-objective hyperparameter selection problem
\begin{equation}
\label{eq:goal}
\begin{aligned}
    &\min_{\lambda \in \Lambda} \; \{ R_{L_c+1}(\lambda), R_{L_c+2}(\lambda), \dots, R_L(\lambda) \} \\
    &\text{subject to} \; R_l(\lambda) < \alpha_l \; \text{for all} \; 1 \leq l \leq L_c,
\end{aligned}
\end{equation}
where the first $L_c\leq L$ risk functions are strictly controlled to be below user-defined thresholds $\{\alpha_l\}_{l = 1}^{L_c}$.

\begin{figure}
    \centering
    \includegraphics[width=\columnwidth]{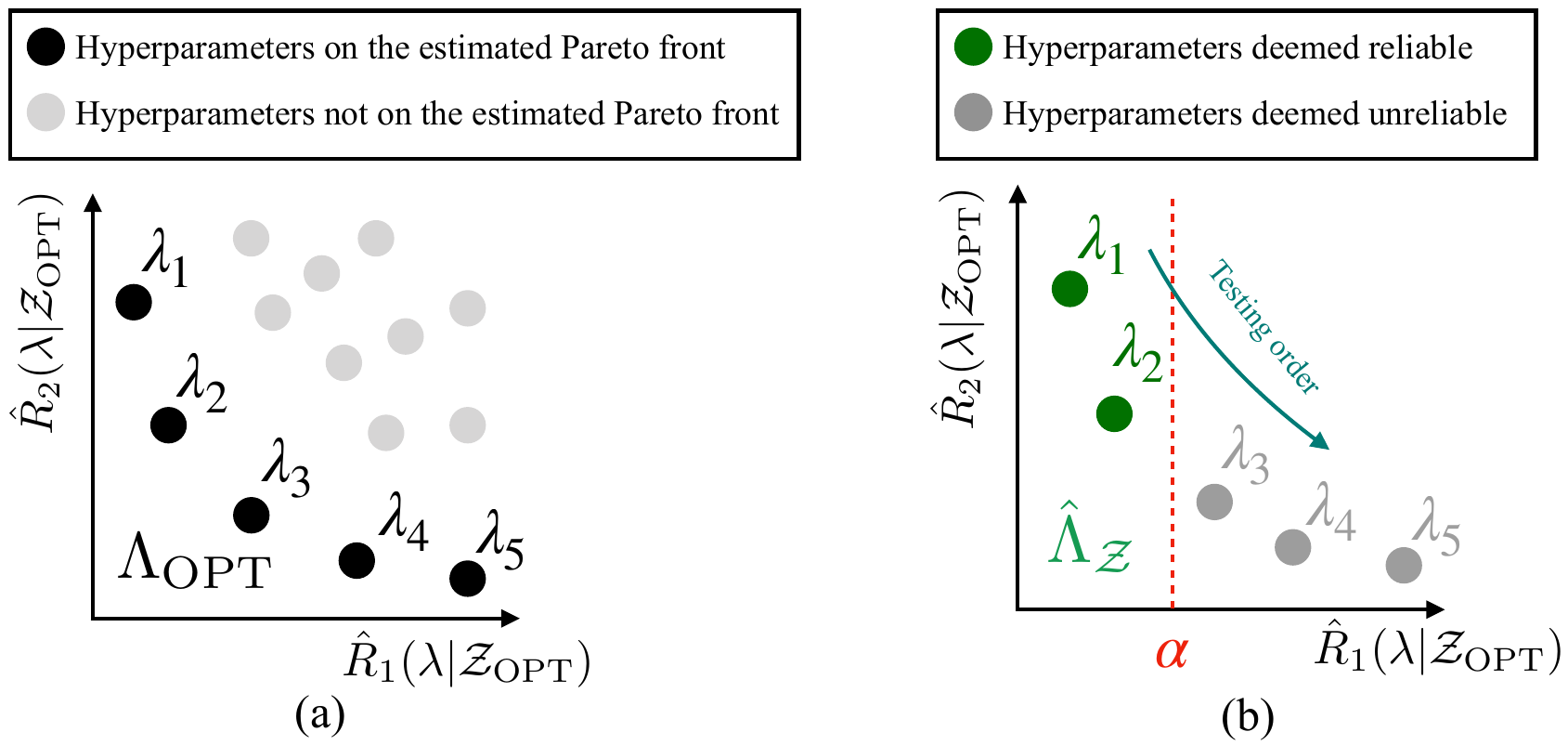}
    \caption{Illustration of the operation of PT for one reliability risk function $R_1(\lambda)$ to be controlled below the threshold $\alpha$ and one auxiliary risk function $R_2(\lambda)$ to be minimized. (a) Find the subset $\Lambda_\text{OPT}$ lying on the estimated Pareto front; (b) Perform MHT via FST to identify the subset $\hat{\Lambda}_\mathcal{Z}$ of reliable hyperparameters.}
    \label{fig:Pareto}
\end{figure}

PT \cite{laufer2022efficiently} begins by partitioning the calibration dataset \(\mathcal{Z}\) into two disjoint subsets \(\mathcal{Z}_\text{OPT}\) and \(\mathcal{Z}_\text{MHT}\). The subset \(\mathcal{Z}_\text{MHT}\) is reserved for the MHT process, while \(\mathcal{Z}_\text{OPT}\) is used for a preliminary optimization step aimed at estimating the Pareto front for the multi-objective problem (\ref{eq:goal}). Specifically, using the dataset \(\mathcal{Z}_\text{OPT}\), PT determines the subset \(\Lambda_\text{OPT} \subseteq \Lambda\) consisting of hyperparameters that form the Pareto front in the space of estimated risk measures \(\{\hat{R}_l^{\text{avg}}(\lambda|\mathcal{Z}_\text{OPT})\}_{l=1}^L\), calculated for each reliability risk function $R_l(\lambda)$ using (\ref{eq:empirical_estimate}). Any standard multi-objective optimization algorithm can be employed for this purpose, as discussed in \cite{laufer2022efficiently}. An illustration of this process is provided in Figure \ref{fig:Pareto}(a).

Given the estimated Pareto front, PT forms the new hypotheses
\begin{equation}
\label{eq:combined_hypothesis}
\mathcal{H}^{\text{multi}}_{\lambda} :\;\text{there exists} \; l \in \{1, \dots, L_c\} \text{ such that } R_l(\lambda) > \alpha_l.
\end{equation}
As such, rejecting the hypothesis $\mathcal{H}^{\text{multi}}_{\lambda}$ is equivalent to detecting $\lambda$ as reliable in terms of the constraint in (\ref{eq:goal}). For the null hypothesis (\ref{eq:combined_hypothesis}), PT computes the combined p-value
\begin{equation}
\label{eq:combined_pval}
p_\lambda(\widetilde{\mathcal{Z}}) = \max_{1 \leq l \leq L_c} p_{\lambda,l}(\widetilde{\mathcal{Z}}),
\end{equation}
where $\{p_{\lambda,l}(\widetilde{\mathcal{Z}})\}_{l = 1}^{L_c}$ are valid p-values for each of the reliability risk functions, and $\widetilde{\mathcal{Z}}$ denotes either $\mathcal{Z}_\text{OPT}$ or $\mathcal{Z}_\text{MHT}$.

To perform MHT, PT first orders the hyperparameters in $\Lambda_\text{OPT}$ based on the p-values $\{p_\lambda(\mathcal{Z}_\text{OPT})\}_{\lambda \in \Lambda_\text{OPT}}$ from low to high to reflect an ordering on their expected reliability, and then uses the p-values $\{p_\lambda(\mathcal{Z}_\text{MHT})\}_{\lambda \in \Lambda_\text{OPT}}$ to perform FST for MHT. This is illustrated in Figure \ref{fig:Pareto}(b).

\section{Side Information}
\label{sec:prior}

In many engineering and real-world applications, we may have access to side information regarding the expected reliability of the hyperparameters in $\Lambda$. 

\begin{figure}
    \centering
    \includegraphics[width=\columnwidth]{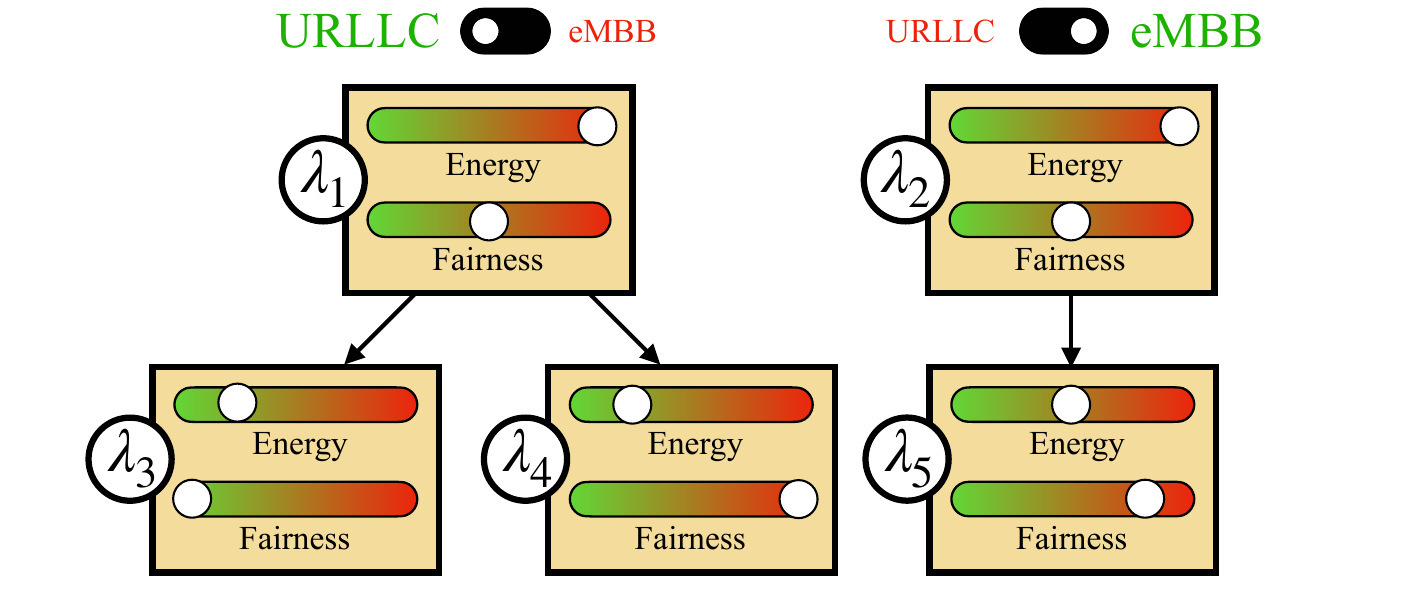}
    \caption{An example of prior information on the relative reliability of hyperparameters. Hyperparameters corresponding to a higher energy consumption are considered more reliable, and put at a higher level in the RG. Only if these configurations are deemed reliable do we proceed to test hyperparameters with better energy efficiency, which may also improve other performance criteria such as fairness.}
    \label{fig:side_info}
\end{figure}

As an example of this, consider a telecommunication network with three interacting AI apps: (\emph{i}) a resource allocation app balancing traffic between enhanced Mobile Broadband (eMBB) services and Ultra-Reliable Low-Latency Communications (URLLC) users \cite{8911618}; (\emph{ii}) a scheduling app enforcing fairness among eMBB users \cite{9878057};  and (\emph{iii}) an energy management app optimizing a base station's energy efficiency \cite{9559261}. In this setting, hyperparameters yielding a larger energy consumption are likely to produce more advantageous KPIs in terms of throughput or latency. As depicted in Figure \ref{fig:side_info}, this information can be encoded using a \textit{reliability graph} (RG).

An RG is a \textit{directed acyclic graph} (DAG) that encodes reliability dependencies among the hyperparameters in \(\Lambda\). In the RG, each hyperparameter in \(\Lambda\) is represented as a node, and a directed edge from \(\lambda_1\) to \(\lambda_2\) indicates that \(\lambda_1\) is expected to be more reliable than \(\lambda_2\). 

\begin{figure}
    \centering
    \includegraphics[width=\columnwidth]{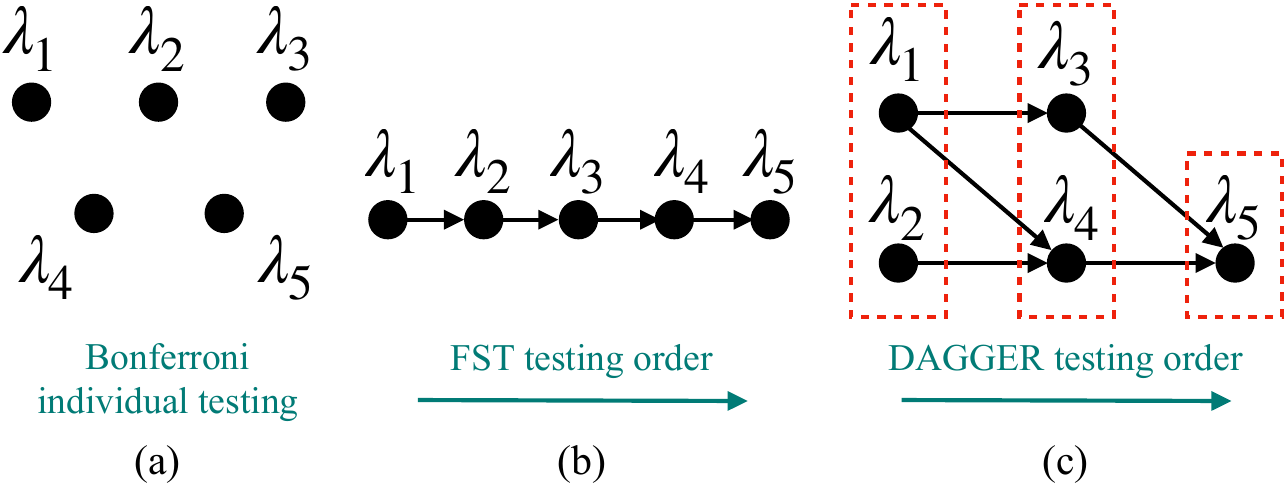}
    \caption{Comparison of (a) Bonferroni method, (b) FST, and (c) DAGGER testing methods.}
    \label{fig:testing_differences}
\end{figure}

RG-PT \cite{farzaneh2025multi} takes as input pairwise probabilities $0\leq \eta_{ij}\leq 1$ for each pair of hyperparameters $\lambda_i, \lambda_j \in \Lambda_\text{OPT}$. The probability $\eta_{ij}$ encodes the prior information on how likely it is for hyperparameter $\lambda_i$ to be more reliable than hyperparameter $\lambda_j$. Additionally, the strength of the prior information is determined via another input $n_p$, which can be seen as a pseudocount variable as in the standard categorical-Dirichlet model \cite{bishop2006pattern}. Using held-out data and this prior knowledge, RG-PT estimates an RG, which is then given as an input to DAGGER \cite{ramdas2017dagger}, an FDR-controlling algorithm for performing MHT on DAGs. As illustrated in Figure \ref{fig:testing_differences}, DAGGER leverages the RG to guide the MHT procedure, and identifies the subset \(\hat{\Lambda}_\mathcal{Z}\) of reliable hyperparameters, ensuring compliance with the FDR constraint in (\ref{eq:FDR_statistical}).

\section{Adaptivity}
\label{sec:adaptive}

The hyperparameter selection schemes considered so far are static, in the sense that they operate with a given fixed held-out data set $\mathcal{Z}$. In contrast, adaptive hyperparameter selection schemes can collect and use data dynamically based on accumulated evidence, potentially reducing the required data while maintaining statistical reliability. This is particularly useful when evaluating risk functions is expensive or risky.

Formally, a test is adaptive if (\textit{i}) it operates sequentially, selecting the next subset of hyperparameters to test based on previous observations; and (\textit{ii}) the testing process can conclude as soon as sufficient evidence has been gathered. 

aLTT \cite{zecchin2024adaptive} is an adaptive LTT-based hyperparameter selection method that improves upon LTT by leveraging sequential data-dependent testing. Building on \cite{ramdas2024hypothesis}, aLTT dynamically selects which hyperparameters to test, and determines when to stop testing. Importantly, aLTT relies on e-values and e-processes, given that p-values are not well suited to support optimal continuation and monitoring \cite{xu2024online}.

In particular, in aLTT, each null hypothesis \( \mathcal{H}_{\lambda_i} : R(\lambda_i) > \alpha \) is tested using the e-value sequence
\begin{equation}
    E_{i,t} = \prod_{t' \leq t} (1 + \mu_{t'} (\alpha - r(\lambda_i,t'))),
\end{equation}
where \( \mu_{t} \) is a betting strategy that adapts to new observations \cite{ramdas2024hypothesis}, and \( r(\lambda_i,t) \) represents the observed risk for hyperparameter \( \lambda_i \) at time \( t \). The use of e-values allows for anytime-valid inference, meaning aLTT can stop early without compromising statistical guarantees \cite{zecchin2024adaptive}.

\section{Conclusion}
\label{sec:conclusion}
In this paper, we have provided a review of LTT-based hyperparameter selection methods, which leverage MHT to ensure the reliability of selected hyperparameters with provable statistical guarantees. We described the fundamental LTT framework and explored several extensions that address critical aspects of hyperparameter selection in various engineering and real-world applications. These include alternative risk measures, multi-objective optimization, the incorporation of prior information, and adaptive testing mechanisms.

The field of hyperparameter selection via MHT is growing quickly, and interesting directions for research include the development of contextual MHT methods and the extension to semi-supervised testing techniques.

\balance


\begin{thebibliography}{10}

\bibitem{angelopoulos2021learn}
A.~N. Angelopoulos, S.~Bates, E.~J. Cand{\`e}s, M.~I. Jordan, and L.~Lei, ``{Learn then test: Calibrating predictive algorithms to achieve risk control},'' {\em arXiv preprint arXiv:2110.01052}, 2021.

\bibitem{farzaneh2024quantile}
A.~Farzaneh, S.~Park, and O.~Simeone, ``Quantile learn-then-test: Quantile-based risk control for hyperparameter optimization,'' {\em IEEE Signal Processing Letters}, 2024.

\bibitem{farzaneh2024statistically}
A.~Farzaneh and O.~Simeone, ``Statistically valid information bottleneck via multiple hypothesis testing,'' {\em arXiv preprint arXiv:2409.07325}, 2024.

\bibitem{laufer2022efficiently}
B.~Laufer-Goldshtein, A.~Fisch, R.~Barzilay, and T.~S. Jaakkola, ``{Efficiently controlling multiple risks with Pareto testing},'' in {\em Proc. International Conference on Learning Representations}, 2023.

\bibitem{farzaneh2025multi}
A.~Farzaneh and O.~Simeone, ``Multi-objective hyperparameter selection via hypothesis testing on reliability graphs,'' {\em arXiv preprint arXiv:2501.13018}, 2025.

\bibitem{zecchin2024adaptive}
M.~Zecchin and O.~Simeone, ``Adaptive learn-then-test: Statistically valid and efficient hyperparameter selection,'' {\em arXiv preprint arXiv:2409.15844}, 2024.

\bibitem{yang2020hyperparameter}
L.~Yang and A.~Shami, ``On hyperparameter optimization of machine learning algorithms: Theory and practice,'' {\em Neurocomputing}, vol.~415, pp.~295--316, 2020.

\bibitem{bergstra2012random}
J.~Bergstra and Y.~Bengio, ``Random search for hyper-parameter optimization.,'' {\em Journal of machine learning research}, vol.~13, no.~2, 2012.

\bibitem{snoek2012practical}
J.~Snoek, H.~Larochelle, and R.~P. Adams, ``Practical bayesian optimization of machine learning algorithms,'' {\em Advances in neural information processing systems}, vol.~25, 2012.

\bibitem{dzau2023achieving}
V.~J. Dzau, M.~H. Laitner, A.~Temple, and T.~H. Nguyen, ``Achieving the promise of artificial intelligence in health and medicine: Building a foundation for the future,'' 2023.

\bibitem{bensalem2023indeed}
S.~Bensalem, C.-H. Cheng, W.~Huang, X.~Huang, C.~Wu, and X.~Zhao, ``What, indeed, is an achievable provable guarantee for learning-enabled safety-critical systems,'' in {\em International Conference on Bridging the Gap between AI and Reality}, pp.~55--76, Springer, 2023.

\bibitem{schuster2022confident}
T.~Schuster, A.~Fisch, J.~Gupta, M.~Dehghani, D.~Bahri, V.~Tran, Y.~Tay, and D.~Metzler, ``Confident adaptive language modeling,'' {\em Advances in Neural Information Processing Systems}, vol.~35, pp.~17456--17472, 2022.

\bibitem{angelopoulos2022image}
A.~N. Angelopoulos, A.~P. Kohli, S.~Bates, M.~Jordan, J.~Malik, T.~Alshaabi, S.~Upadhyayula, and Y.~Romano, ``Image-to-image regression with distribution-free uncertainty quantification and applications in imaging,'' in {\em International Conference on Machine Learning}, pp.~717--730, PMLR, 2022.

\bibitem{karasik2022learning}
R.~Karasik, O.~Simeone, H.~Jang, and S.~S. Shitz, ``Learning to broadcast for ultra-reliable communication with differential quality of service via the conditional value at risk,'' {\em IEEE Transactions on Communications}, vol.~70, no.~12, pp.~8060--8074, 2022.

\bibitem{sun2015multi}
Y.~Sun, D.~W.~K. Ng, and R.~Schober, ``Multi-objective optimization for power efficient full-duplex wireless communication systems,'' in {\em 2015 IEEE Global Communications Conference (GLOBECOM)}, pp.~1--6, IEEE, 2015.

\bibitem{raviv2024adaptive}
T.~Raviv, S.~Park, O.~Simeone, Y.~C. Eldar, and N.~Shlezinger, ``Adaptive and flexible model-based ai for deep receivers in dynamic channels,'' {\em IEEE Wireless Communications}, 2024.

\bibitem{hou2025if}
Q.~Hou, S.~Park, M.~Zecchin, Y.~Cai, G.~Yu, and O.~Simeone, ``What if we had used a different app? reliable counterfactual kpi analysis in wireless systems,'' {\em IEEE Transactions on Cognitive Communications and Networking}, 2025.

\bibitem{ramdas2024hypothesis}
A.~Ramdas and R.~Wang, ``Hypothesis testing with e-values,'' {\em arXiv preprint arXiv:2410.23614}, 2024.

\bibitem{rice2007mathematical}
J.~A. Rice, {\em Mathematical Statistics and Data Analysis.}
\newblock Belmont, CA: Duxbury Press., third~ed., 2006.

\bibitem{abdar2021review}
M.~Abdar, F.~Pourpanah, S.~Hussain, D.~Rezazadegan, L.~Liu, M.~Ghavamzadeh, P.~Fieguth, X.~Cao, A.~Khosravi, U.~R. Acharya, {\em et~al.}, ``A review of uncertainty quantification in deep learning: Techniques, applications and challenges,'' {\em Information fusion}, vol.~76, pp.~243--297, 2021.

\bibitem{rodriguez2024information}
B.~Rodr{\'\i}guez-G{\'a}lvez, R.~Thobaben, and M.~Skoglund, ``An information-theoretic approach to generalization theory,'' {\em arXiv preprint arXiv:2408.13275}, 2024.

\bibitem{Nokia}
A.~Valcarce, ``{Wireless Suite: A collection of problems in wireless telecommunications},'' {\em https://github.com/nokia/wireless-suite}, 2020.

\bibitem{tishby2000information}
N.~Tishby, F.~Pereira, and W.~Bialek, ``The information bottleneck method,'' in {\em Proc. Allerton Conference on Communication, Control and Computation}, 2001.

\bibitem{bauer1991multiple}
P.~Bauer, ``Multiple testing in clinical trials,'' {\em Statistics in Medicine}, vol.~10, no.~6, pp.~871--890, 1991.

\bibitem{einbinder2024semi}
B.-S. Einbinder, L.~Ringel, and Y.~Romano, ``Semi-supervised risk control via prediction-powered inference,'' {\em arXiv preprint arXiv:2412.11174}, 2024.

\bibitem{benjamini1995controlling}
Y.~Benjamini and Y.~Hochberg, ``Controlling the false discovery rate: a practical and powerful approach to multiple testing,'' {\em Journal of the Royal statistical society: series B (Methodological)}, vol.~57, no.~1, pp.~289--300, 1995.

\bibitem{8911618}
M.~Elsayed and M.~Erol-Kantarci, ``{AI-Enabled Radio Resource Allocation in 5G for URLLC and eMBB Users},'' in {\em 2019 IEEE 2nd 5G World Forum (5GWF)}, pp.~590--595, 2019.

\bibitem{9878057}
X.~Han, K.~Xiao, R.~Liu, X.~Liu, G.~C. Alexandropoulos, and S.~Jin, ``Dynamic resource allocation schemes for embb and urllc services in 5g wireless networks,'' {\em Intelligent and Converged Networks}, vol.~3, no.~2, pp.~145--160, 2022.

\bibitem{9559261}
T.~Rumeng, W.~Tong, S.~Ying, and H.~Yanpu, ``Intelligent energy saving solution of 5g base station based on artificial intelligence technologies,'' in {\em 2021 IEEE International Joint EMC/SI/PI and EMC Europe Symposium}, pp.~739--742, 2021.

\bibitem{bishop2006pattern}
C.~M. Bishop and N.~M. Nasrabadi, {\em Pattern recognition and machine learning}, vol.~4.
\newblock Springer, 2006.

\bibitem{ramdas2017dagger}
A.~Ramdas, J.~Chen, M.~J. Wainwright, and M.~I. Jordan, ``A sequential algorithm for false discovery rate control on directed acyclic graphs,'' {\em Biometrika}, vol.~106, no.~1, pp.~69--86, 2019.

\bibitem{xu2024online}
Z.~Xu and A.~Ramdas, ``Online multiple testing with e-values,'' in {\em International Conference on Artificial Intelligence and Statistics}, pp.~3997--4005, PMLR, 2024.

\end{thebibliography}
\end{document}